# Efficient Inference on Generalized Fault Diagrams

Ross D. Shachter and Leonard J. Bertrand
Department of Engineering-Economic Systems, Stanford University
(visiting the Center for Health Policy Research and Education,
Duke University, PO Box GM, Durham, NC 27706)
and Strategic Decisions Group, Menlo Park, CA

The generalized fault diagram, a data structure for failure analysis based on the influence diagram, is defined. Unlike the fault tree, this structure allows for dependence among the basic events and replicated logical elements. A heuristic procedure is developed for efficient processing of these structures.

Deterministic logic and conditional probabilities are both appealing frameworks in which to build a knowledge base. Each has a natural graphical representation, semantic network for logic and influence diagrams (Howard and Matheson, 1981) or bayes networks (Pearl, 1986) for probabilities. Deterministic logic lends itself to efficient inference over large knowledge bases, either goal- or data-driven. Probabilities, on the other hand, permit the knowledge engineer to express information about uncertainty. Unfortunately, this often comes at the expense of efficient analysis. A desirable goal is a combination of the two frameworks that exploits the efficiency of deterministic logic but allows the richness of uncertain elements.

The classic fault tree is an example of a problem structure with a deterministic logical skeleton and underlying uncertain components. Because of the rigid assumptions of conditional independence built into the fault tree, it lends itself to straightforward analysis in linear time. This is easily seen in the influence diagram representation or the work of Pearl (1986) on singly connected graphs. For most problems, however, the fault tree is too restrictive. It is unable to recognize the dependence among uncertain components and, at a higher level in the hierarchy, the dependence of a single subsystem on multiple systems.

A new structure, the fault influence diagram, is developed, which generalizes the fault tree by explicitly relaxing many of the conditional independence assumptions. In this framework, there may be explicit dependence among probabilistic components and subsystems may contribute

413

to the success (or failure) of multiple systems. Since these wrinkles add multiple paths from the basic components to the top event, they can significantly increase the complexity of the analysis.

There are three types of problems which can be considered in this framework. Previously, we have considered the general probabilistic problem, and how to obtain a full conditional distribution for a variable of interest conditioned on all possible values of variables to be observed. Pearl (1986) has examined similar networks in the presence of particular observations. In this present work based on Bertrand (1986), we restrict ourselves to the unconditional distribution for a variable of interest, so that we can concentrate on efficient evaluation. Further research could extend our results to the conditional cases.

### The Fault Influence Diagram

Our representation for the generalized fault tree is an influence diagram which recognizes logical relations. We assume that all events are binary, either success or failure, and that each variable must have a distribution conditioned on its immediate predecessors in the graph. (In our representation the top event is a sink in the network, and will be the variable of interest.) A distribution can be either deterministic, via the logical operators AND, OR, and NOT, or probabilistic, in the form of a conditional probability distribution. If the logical operator were "probabilized", that is if they were represented explicitly, it would be a full conditional distribution, a general truth table. Thus, one key to efficient processing is the "isolation" of logic from probabilities, so that the logical operator remain in their implicit form. (For example, if any of the predecessor events for an AND is known to have failed, then the AND has also failed, and this is easy to recognize. However, if the AND were represented by a general conditional probability distribution, this would be much harder–and not very practical–to recognize.)

In a fault influence diagram, the probability of success for a logical operator may depend on other logical operators and probabilistic events. However, to simplify our analysis, we require that the probability distribution of a probabilistic event can depend only on other probabilistic variables.

A full fault influence diagram that will be used to demonstrate our solution procedure, is depicted in figure 1. Every node in the diagram corresponds to a variable. Logical operators are labelled as such. Each node with an unconditional probability distribution has its probability of success indicated. For the sake of simplicity, conditional distributions are not shown.

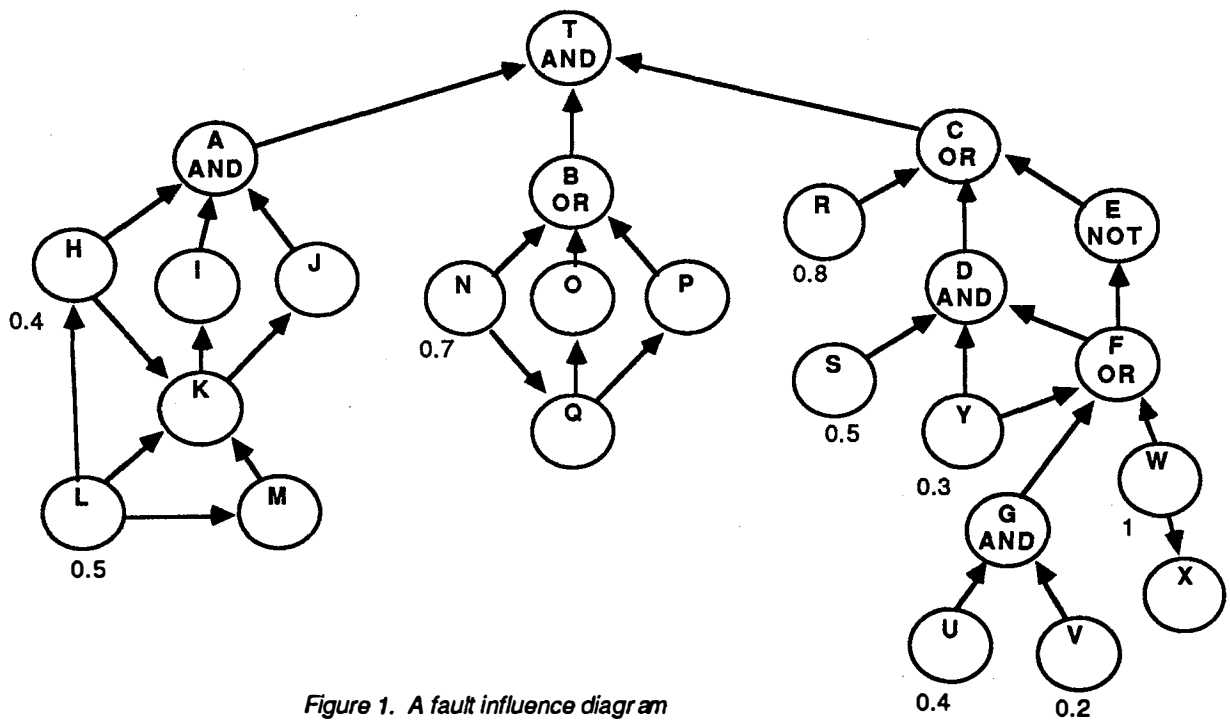

*Figure 1. A fault influence diagram*

## The Procedure

Our procedure determines the probability of success of the top event by successively reducing nodes or blocks of nodes in the graph. This can be done using the influence diagram manipulations (Shachter, 1986a) of arc reversal (Bayes' Theorem) and node removal (conditional expectation), but these operations do not take advantage of the logical structure. A more appealing method used by both Pearl (1986) and McCullers (1985) is to instantiate on the possible values of a set of variables, so as to render the problem (or the module within it) singly connected. We can then condition the results by the probability of each instantiation. For this process to succeed, we must make a good choice of variable to instantiate, and this is the kernel of our method.

The procedure starts by pre-processing the fault diagram. Events with no directed path from them to the top node, such as node X in figure 1, do not bear on the top event. Such events can be easily recognized and eliminated from the analysis (Shachter, 1986a). Likewise, events can be <u>trimmed</u> if they are known for certain (with probability one or zero), and singly connected subbranches can be reduced. In figure 1, node W is eliminated and ensures that node F, being an OR operator, will succeed. The module formed by G, U , and V can be eliminated and the arc between Y and F deleted. Since F succeeds with probability one, the arc between F and D can be deleted, with the probability of D still depending on S and Y, and E, being a NOT operator, is assigned a zero probability. Finally, E and the arc between E and C are deleted.

The next stage in the algorithm, <u>computing</u>, involves the simple computation of probabilities for logical operators. It can be performed for any logical operator with singly-connected probabilistic predecessors. Computing involves replacing a logical operator with a probabilistic node in the fault diagram. The probability of success of an AND operator is the product of the probability of success of its predecessors. For an OR operator, the probability of success is one minus the product of the probability of failure of its predecessors. For a NOT, its probability of success is the probability of failure of its predecessor. In the example, computing will first reduce node D and then node C.

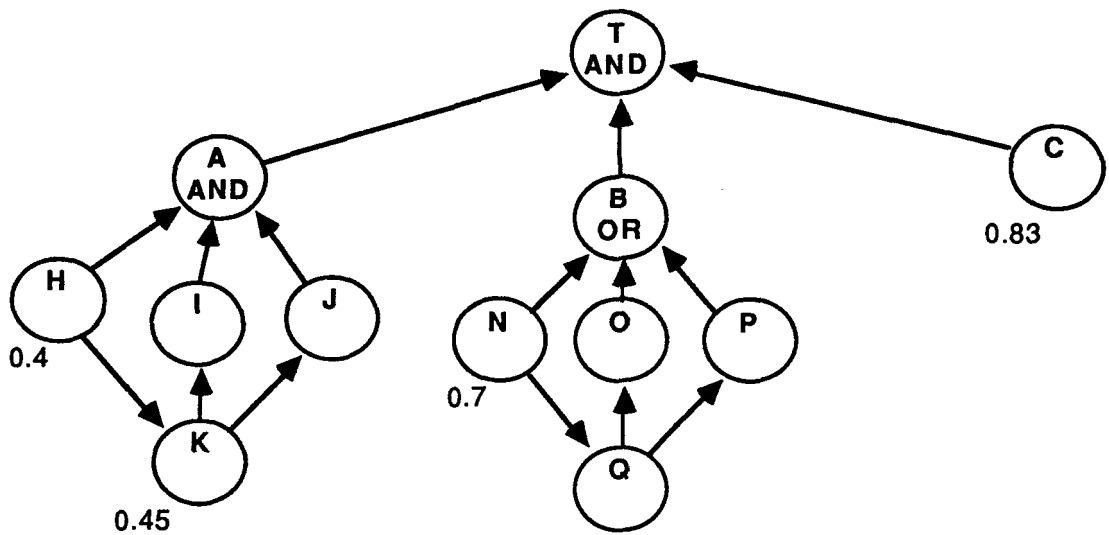

*Figure 2. Fault influence diagram after pre-processing step.*

After these trimming and computing steps, influence diagram manipulations can be used to reduce additional nodes from the graph. Any probabilistic node with a single, probabilistic successor can be removed without "probabilizing" logical elements. In this way, node M, in figure 1, can be removed into node K. There are some other nodes which can also be reduced through a combination of arc reversal and node removal (Shachter, 1986b). Such nodes, called <u>grandfathers,</u> are defined as probabilistic events which are more than two arcs from a logical operator along every directed path. Again, in figure 1, node L is a grandfather and is reduced by first reversing the arc between L and H and then removing it into K. Figure 2 shows the fault influence diagram after the pre-processing step.

The procedure next identifies <u>partitions</u>, either contiguous sets of probabilistic nodes or logical events with multiple successors. In figure 2, we can identify three partitions: {H, I, J, K}, {N, O, P, Q}, and {C}. For each partition that is not simply a single probabilistic event, we must find the "immediate reverse dominator" (IRD), the closest node that is on every path from the partition to the top event. A <u>control graph</u> can now be constructed: the nodes are the partitions and the arcs indicate the connections among the partitions. The control graph for the example in figure 2 is shown in figure 3.

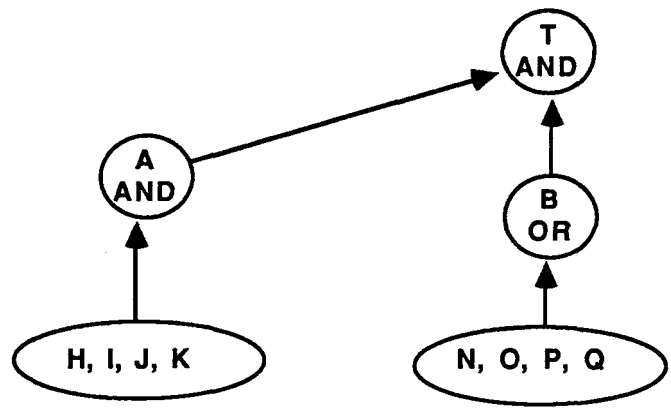

Figure 3. Control graph showing partitions and their IRDs.

Given a cut-vertex in the diagram, a <u>module</u> is defined as the cut-vertex plus all nodes which can reach it, that is, all of the nodes which would be disconnected from the top event if the cut-vertex were deleted. A module is just a smaller fault diagram. Using the control graph, we can heuristically determine which module to process next and the order in which to instantiate the partitions, if there are multiple partitions within the module. To choose which module to process next, we select the partition "closest" to the top node in the control graph. If there is more than one such partition, select the one that has the most outgoing arcs. In our example, both partitions tie for both criteria, so we will arbitrarily choose to process the partition {N, O, P, Q}. Once we have selected a partition, we then choose that module corresponding to the cut-vertex that is closest to the IRD of the partition. In our example, we should process the module with cut vertex B.

Now that we have chosen which subproblem should be analyzed, we have to identify which variable should be instantiated to obtain a singly-connected graph. Another type of graph, the <u>partition graph</u>, is constructed. This graph consists of the nodes in the partition and the arcs between variables in the partition. A sink node (*) is added and receives an arc from any variable in the partition with a logical successor. In the partition graph, influence diagram operations will be simulated. The heuristic involves perturbing the dependencies (reversing arcs) within the partition to determine which nodes can simply be integrated out of the module and which nodes must be instantiated. A <u>candidate source node</u> (CSN) is a probabilistic node with at least one probabilistic successor. The CSNs are the source of dependency in the module. We identify all CSNs and inspect them in order from sink to source. At any point, if a CSN has only logical successors, then it will require neither instantiation nor reduction. (It will be reduced through computing.) If it has exactly one probabilistic successor then it can be reduced through influence diagram operations. Otherwise it will need to be instantiated. When such a node is encountered, all arcs into it should be reversed, so other nodes might avoid instantiation. Figure 4 demonstrates this process. starting with the partition graph, nodes N and Q are identified as CSN. We visit Q first, realize that it will require instantiation, and reverse the arc into it. Now we find that N will not need to be instantiated.

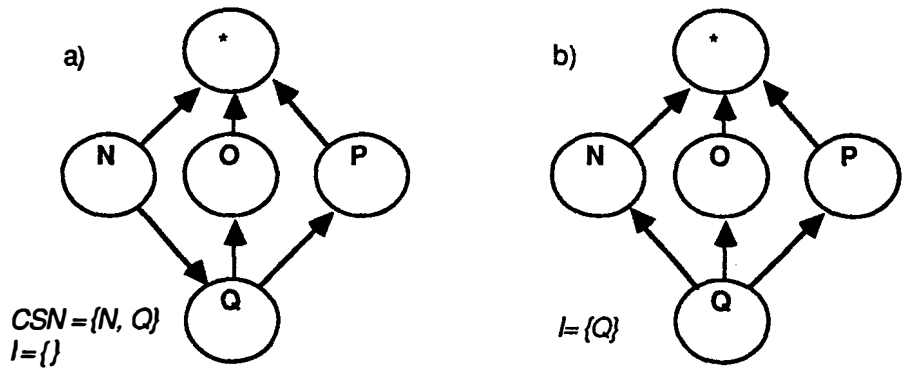

Figure 4. Finding the variables to instantiate. Partition graphs (a) before and (b) after inspection.

The next step in our algorithm is to proceed with the instantiation of Q. Two modules are then created similar to the one in figure 5, one conditioned on the success of Q and the other on its failure, and the conditional probabilities for each case are attached to each node. Each instantiated module is solved and the probability of the top node of the module (the cut-vertex) is obtained by expectation over the two possible values that Q can obtain. At this point the complete module corresponding to B is replaced by a probabilistic node with the calculated probability as illustrated in figure 6.

418

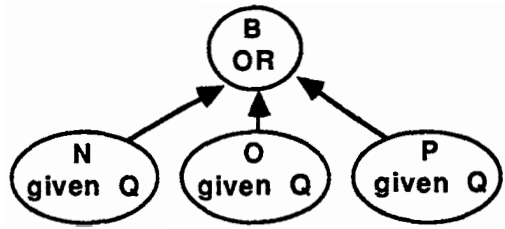

Figure 5. Instantiated module

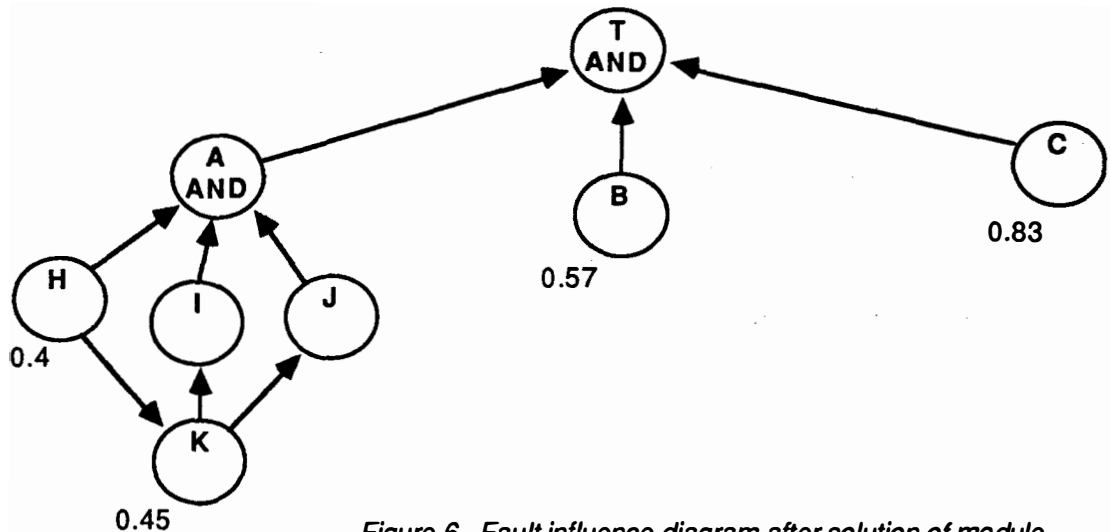

Figure 6. Fault influence diagram after solution of module

At this point, if it is possible, we should trim and compute to further reduce the graph. If we have been able to reduce the diagram to a single probabilistic node, then we have finished. Otherwise, we should proceed to instantiate another node in the partition, if there are any left to be instantiated. If the partition has been completely reduced, then a new partition is selected and the algorithm repeats on the appropriate module. In our example, we would next consider the partition {H, I, J, K} and solve the module corresponding to A. In this case, only node K needs to be Instantiated, and the module can be solved to obtain the diagram in figure 7. The solution is then completed by computing the top event T.

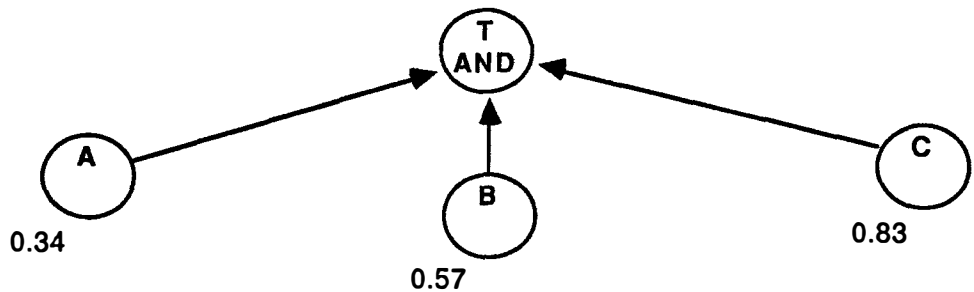

Figure 7. Fault influence diagram after solution of second module.

## Conclusions

The algorithm sketched above is a hybrid approach to the solution of mixed logical-probabilistic inference problems. By recognizing the partitions within the graph and their relationship, and by manipulating the variables within partitions, we are able to reduce dimension of instantiations needed to process the diagram while maintaining the logical structure. This heuristic algorithm is not the "final solution" for problem of this type, but it is a substantial improvement over the existing techniques.